\documentclass[conference]{IEEEtran}
\IEEEoverridecommandlockouts
\usepackage{cite}
\usepackage{amsmath,amssymb,amsfonts}
\usepackage{algorithmic}
\usepackage{graphicx}
\usepackage{textcomp}

\usepackage[table]{xcolor}
\usepackage{colortbl}
\usepackage{booktabs, makecell}
\usepackage[pagebackref=false,breaklinks=true,colorlinks,bookmarks=false]{hyperref}
\usepackage{cleveref}
\usepackage[accsupp]{axessibility}
\usepackage{nicematrix}
\usepackage[hang,flushmargin]{footmisc}
\usepackage{multirow}

\renewcommand{\baselinestretch}{0.975}

\def\BibTeX{{\rm B\kern-.05em{\sc i\kern-.025em b}\kern-.08em
    T\kern-.1667em\lower.7ex\hbox{E}\kern-.125emX}}

\usepackage{fancyhdr}
\fancypagestyle{firststyle}
{
   \fancyhf{}
   \fancyhead[C]{
    \textcolor{red}{
       The content of this paper was presented at IWBF 2024. If you find this research useful, please cite the following paper: \\ Ž. Babnik, F. Boutros, N. Damer, P. Peer, V. Štruc, "AI-KD: Towards Alignment Invariant Face Image Quality Assessment Using Knowledge Distillation" in Proceedings of the IEEE International Workshop on Biometrics and Forensics (IWBF), 2024.}
    }
}

\begin{document}

\title{AI-KD: Towards Alignment Invariant Face Image Quality Assessment Using Knowledge Distillation \vspace{-1mm}
\thanks{\scriptsize Supported in parts by the ARIS junior researcher program, ARIS Programmes P2-0250 ``Metrology and Biometric Systems'' and P2-0214 ``Computer Vision'', and ARIS Research Project J2-50069 ``MIXBAI'', as well as by the German Federal Ministry of Education and Research and the Hessian Ministry of Higher Education, Research, Science and the Arts within their joint support of the National Research Center for Applied Cybersecurity ATHENE.}
}

\author{\IEEEauthorblockN{Žiga Babnik$^1$, Fadi Boutros$^2$, Naser Damer$^{2,3}$, Peter Peer$^1$, and Vitomir Štruc$^1$}
\IEEEauthorblockA{$^1$University of Ljubljana, Ljubljana, Slovenia}
\IEEEauthorblockA{$^2$Fraunhofer Institute for Computer Graphics Research IGD, Darmstadt, Germany}
\IEEEauthorblockA{$^3$Department of Computer Science, TU Darmstadt, Darmstadt, Germany}
{\small \url{https://github.com/LSIbabnikz/AI-KD}}
}



\IEEEoverridecommandlockouts \IEEEpubid{\makebox[\columnwidth]{979-8-3503-5447-8/24/\$31.00 ©2024 IEEE \hfill} \hspace{\columnsep}\makebox[\columnwidth]{ }}

\maketitle
\thispagestyle{firststyle}


\begin{abstract}
Face Image Quality Assessment (FIQA) techniques have seen steady improvements over recent years, but their performance still deteriorates if the input face samples are not properly aligned. This alignment sensitivity comes from the fact that most FIQA techniques are trained or designed using a specific face alignment procedure. If the alignment technique changes, the performance of most existing FIQA techniques quickly becomes suboptimal. To address this problem, we present in this paper a novel knowledge distillation approach, termed AI-KD that can extend on any existing FIQA technique, improving its robustness to alignment variations and, in turn, performance with different alignment procedures. To validate the proposed distillation approach, we conduct comprehensive experiments on $6$ face datasets with $4$ recent face recognition models and in comparison to $7$ state-of-the-art FIQA techniques. Our results show that AI-KD consistently improves performance of the initial FIQA techniques not only with misaligned samples, but also with properly aligned facial images. Furthermore, it leads to  a new state-of-the-art, when used with a competitive initial FIQA approach. The code for AI-KD is made publicly available from: {\small \url{https://github.com/LSIbabnikz/AI-KD}}.
\end{abstract}


\section{Introduction}\label{sec:introduction}

Face Image Quality Assessment~(FIQA) refers to the process of predicting quality scores for facial images, which can be used to control the biometric capture process and to provide feedback either to the subject or to an automated face recognition (FR) system. 
This is particularly important for unconstrained image acquisition scenarios, where sample quality can vary significantly and can, therefore, have an adverse impact on performance~\cite{fr1,survey}. 

Modern FIQA approaches typically predict a single numerical value from the input face samples, also referred to as a \textit{unified quality score}, that aims to capture the biometric utility of the given sample for the recognition task~\cite{eval2}. While significant progress has been made in FIQA techniques over the years \cite{survey}, 
existing techniques are still sensitive to the alignment of the input samples~\cite{fr_align}. 
The main reason for this sensitivity is that most FIQA techniques are trained on samples aligned using a specific facial landmark detector (also often referred to as a face keypoint detector), and, as such, also often overfit to that particular landmark detector. 
Even though modern landmark detectors are robust and perform well on challenging benchmarks~\cite{face_det}, using an unknown detector that was not seen during training, still leads to a notable decrease in FIQA performance. 

To address this issue, we present in this paper an \textbf{A}lignment-\textbf{I}nvariant \textbf{K}nowledge \textbf{D}istillation (AI-KD) procedure that improves the performance of existing FIQA approaches, when dealing with samples produced using any (unknown) face landmark detector. AI-KD relies on a novel distillation process that incorporates simple image transformations, which mimic the (minor) variability between samples produced by different alignment approaches. Using multiple FIQA techniques, FR models and performance benchmarks, we show that AI-KD leads to considerable performance gains when confronted with misaligned face images, but also improves performance when the face samples are optimally aligned.   

\begin{figure*}[!ht]
    \centering
    \includegraphics[width=0.80\textwidth]{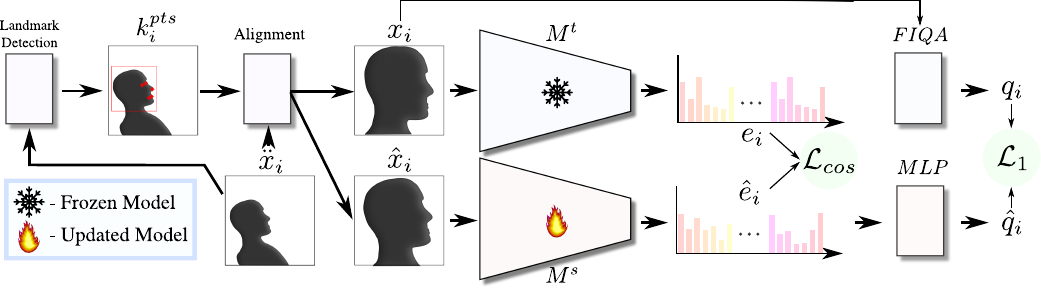}
    \caption{\textbf{Overview of the proposed Alignment Invariant Knowledge Distillation (AI-KD) process.} The proposed approach trains a quality-regression model, consisting of a FR backbone $M^s$ and a quality regression head $MLP$, on quality labels $q_i$ extracted using any existing FIQA approach. Training samples $\hat{x}_i$ are (mis)aligned on the fly, by perturbing the correct landmark $k^{pts}_i$ of the initial unaligned face samples $\ddot{x}_i$. Additionally, to ensure robustness to alignment variations in the distilled model, we design a distillation objective that ensures consistency between representations of the aligned $e_i$ and (mis)aligned images $\hat{e}_i$, as well as matching the predicted quality scores $\hat{q}_i$ to the quality labels $q_i$. }
    \label{fig:method}
\end{figure*}

\section{Related Work}\label{sec:related_work}


\subsection{General FIQA Techniques}\label{sec:related_work:indirect_fiqa}

General FIQA techniques assess the (biometric) quality of the given face samples, by predicting a single (unified) numerical score, where a higher score usually refers to a higher quality. 
Based on how the methods compute the quality score, they can be further divided into: unsupervised (analytical) methods and supervised (regression-based) methods.

\vspace{1mm}\noindent\textbf{Unsupervised methods} estimate the quality of the provided face samples by analyzing their characteristics. Early techniques observed (human) perceptual characteristics, such as noise levels, blurriness, lighting conditions, etc.~\cite{analy11, analy12}. More recent techniques, on the other hand, analyze the image characteristics from the viewpoint of a FR model. These methods most often measure the robustness of the sample's representation in the embedding space of the targeted FR model. 
The earliest such method, named SER-FIQ~\cite{serfiq}, uses Dropout to produce several representations of a single input sample, whereas FaceQAN~\cite{faceqan} uses adversarial methods to generate a number of perturbed samples that are then utilized for quality estimation. The most recent, DifFIQA~\cite{diffiqar} approach, uses the forward and backward processes of diffusion models to generated perturbations of the face samples.
 
\vspace{1mm}\noindent\textbf{Supervised methods} commonly train a quality-regression model to assess the quality of the studied face samples. The regression models are trained on pseudo-quality labels obtained with various strategies. Early works relied on visual image characteristics, similar to unsupervised methods, or human annotations~\cite{bestrowden} to compile the labels. 
Conversely, modern approaches consider FR models and the recognition procedure in the label generation process. FaceQnet~\cite{faceqnet}, for example, compares all images of an individual to its highest quality sample. A similar approach, PCNet~\cite{pcnet}, compares several pairs of images of the same individual to generate labels, while a more advanced approach SDD-FIQA~\cite{sddfiqa} takes into account also the imposter pairs, or the pairs containing two images of distinct individuals when producing pseudo-quality labels for training.

\subsection{Quality-Aware FR Techniques}\label{sec:related_work:direct_fiqa}

Unlike general FIQA techniques, quality-aware FR techniques combine  quality estimation and face recognition into a single task. Here, FR models are trained to discern the identity as well as the quality of the input face samples, by employing standard training procedures combined with a quality regression branch. One of the earliest examples of such methods is PFE~\cite{pfe}, which measured the uncertainty of the samples representation in the latent space of the recognition model. Uncertainty, in this case, can be viewed as the inverse measure of quality. The MagFace~\cite{magface} approach, extends the ArcFace~\cite{arcface} margin-loss with a magnitude-aware term, allowing it to encode quality into the magnitude of the sample representations. The most recent approach, CR-FIQA \cite{crfiqa}, estimates quality as the ratio between the distance of a sample representation to the positive class-center and the nearest negative-class center.

\section{Methodology}\label{sec:methodology}


Face Image Quality Assessment (FIQA) techniques require properly aligned input samples in order to achieve the best possible performance, consequently limiting the choice of landmark detection algorithms used during inference. Although most landmark detectors achieve high accuracy, their predictions usually differ by several pixels, which is enough to adversely impact the performance of any existing FIQA technique. In this paper, we present a simple, but elegant knowledge distillation approach named AI-KD (\textbf{A}lignment-\textbf{I}nvariant \textbf{K}nowledge \textbf{D}istillation), which aims to improve the robustness and overall performance of FIQA techniques when dealing with samples aligned using different landmark detectors. The method, presented in Fig.~\ref{fig:method}, uses knowledge distillation to fine-tune a preexisting FR model $M$ to predict quality scores $\hat{q}_i$ of the input samples $\hat{x}_i$, through an additional $MLP$ (Multi Layer Perceptron) quality-regression head. During training two copies of $M$ are utilized, i.e., the teacher $M^t$ (with frozen weights) is used to generate ground-truth representations of properly aligned input samples $x_i$, while the student $M^s$ is trained to be robust to alignment variations by learning from improperly aligned samples $\hat{x}_i$. In the following sections, we present the whole AI-KD methodology in-detail.

\subsection{Method Overview}\label{sec:methodology:overview}

The proposed AI-KD technique aims to improve the performance of any existing FIQA method, when presented with input samples aligned with an unknown landmark detection algorithm. Formally, given a FIQA technique $Q$, the goal is to train a quality-regression model, consisting of a pretrained FR backbone $M^s$ and an $MLP$ based quality-regression head, on a large face dataset $\{\ddot{x}_i\}_{i=0}^N$ containing $N$ (in general non-aligned) samples. Towards this end, we first extract facial landmarks $k^{pts}_i$ and pseudo quality labels $q_{i}$ from all samples in the datasets. During training, we then employ the alignment-invariant knowledge distillation procedure, which dynamically transforms each initial face sample $\ddot{x}_i$ into a properly aligned sample $x_i$ and a misaligned sample $\hat{x}_i$, imitating alignment variations of different landmark detectors. The knowledge distillation process uses both, a quality $\mathcal{L}_1$ and feature loss $\mathcal{L}_{cos}$, to simultaneously ensure optimal FIQA performance and alignment-invariance of the final trained/distilled model. 

\subsection{Data Preprocessing}\label{sec:methodology:preprocessing}

During the preprocessing step, the facial landmarks $k^{pts}_i$ of all $N$ samples $\ddot{x}_i$ are extracted using a chosen landmark detector $D$, such that $k^{pts}_i = D(\ddot{x}_i)$. The extracted landmarks $k^{pts}_i$, consist of the coordinates of the left and right eye, the tip of the nose and the corners of the lips, and can be used to properly align $\ddot{x}_i$ by matching the coordinates to a predefined template, resulting in a properly aligned sample $x_i$ \cite{arcface,crfiqa,adaface}. Additionally,  pseudo quality labels $q_i$  are also calculated for all $N$ samples $\ddot{x}_i$ using a chosen FIQA technique with the well-aligned samples $q_i = Q(x_i)$.

\subsection{Alignment-Invariant Knowledge Distillation}\label{sec:methodology:knowledge_distillation}

The knowledge distillation procedure uses the landmarks $k^{pts}_i$ and pseudo-quality labels $q_i$ extracted in the data preprocessing step to train a quality-regression model, consisting of a pretrained FR backbone and a quality regression head, i.e., $M^s\circ MLP$. The training process consists of two main steps: $(i)$ the sample transformation step and $(ii)$  the actual knowledge distillation. The sample transformation step generates samples with varying alignments, whereas the knowledge distillation step transfers the knowledge encoded in the pseudo-quality labels to the student model.

\vspace{1mm}\noindent\textbf{Sample Transformation Step.} During this step, the initial face sample $\ddot{x}_i$ is used to generate a properly aligned sample $x_i$ and a misaligned sample $\hat{x}_i$ using the extracted landmarks $k^{pts}_i$. The properly aligned sample $x_i$ is generated by aligning according to $k^{pts}_i$, while $\hat{x}_i$ aims to replicate the alignment produced by an unknown facial landmark detector. Since it is not feasible to extract landmarks of all samples $x_i$ using a large number of unique landmark detectors, we make a simple assumption that the predicted landmarks of any well-functioning face landmark detection method will be approximately similar to the baseline landmarks $k^{pts}_i$. Using this assumption, we then generate new landmarks $\hat{k}^{pts}_i$ corresponding to an unknown method $\hat{D}$, by randomly sampling around the reference coordinates in $k^{pts}_i$. 
Formally, this can be written as: 
\begin{equation}
    \hat{k}^{pts}_i = k^{pts}_i + \mathcal{U}_{[-p,p]},
\end{equation}
where $\mathcal{U}_{[-p,p]}$ is a uniform random variable sampled from the interval $[-p, p]$. This means that all coordinates can differ at most by $p$ pixels between the two landmarks (initial and perturbed). The misaligned sample $\hat{x}_i$ is then produced by aligning $\ddot{x}$ using the newly constructed landmarks $\hat{k}^{pts}_i$.

\vspace{1mm}\noindent\textbf{Knowledge Distillation.} The $N$ training samples $x_i$ and $\hat{x}_i$ produced by the sample transformation step form the basis for the knowledge distillation procedure. Here, the teacher model $M^t$ is frozen during the entire training process (its weights are not updated), while the parameters of the student model $M^s\circ MLP$ are optimized using dedicated distillation objectives. The properly aligned sample $x_i$ is fed through the frozen teacher model $M^t$, to produce a feature representation $e_i = M^t(x_i)$ of the input sample. The computed representation $e_i$ and the corresponding pseudo-quality label $q_i$ jointly represent the regression targets for the student model $(M^s \circ MLP)$. To be able to define a loss for the distillation procedure, the misaligned sample $\hat{x}_i$ is  fed through $M^s$, producing the feature representation of the misaligned sample $\hat{e}_i$. The computed representation $\hat{e}_i$ is then further processed through an $MLP$ to produce the predicted quality label $\hat{q}_i$. The overall distillation objective used to optimize $M^s \circ MLP$ is then defined as the average of a representation and a quality term. Here, the representation loss, aims to align the representations of $\hat{x}_i$ and $x$:
\begin{equation}
    \mathcal{L}_{cos}(e_i, \hat{e_i}) = 1 - \frac{e_i \cdot \hat{e}_i^T}{\|e_i\|\cdot\|\hat{e}_i\|},
\end{equation}
where $\|e_i\|$ represents the norm of the representation, while the quality loss aims to ensure that the original and predicted quality scores are as close as possible, i.e.: 
\begin{equation}
    \mathcal{L}_1(q_i, \hat{q}_i) = |q_i - \hat{q}_i|.
\end{equation}


\begin{table}[!t!]
    \centering
    \renewcommand{\arraystretch}{1.0}
    \caption{Summary of the characteristics of the experimental datasets.}
\resizebox{0.9\columnwidth}{!}{%
    \begin{tabular}{l l c c c c c }
        \toprule
        \multirow{ 2}{*}{\textbf{Dataset}} & \multirow{ 2}{*}{\textbf{\#Images}} & \multirow{ 2}{*}{\textbf{\#IDs}} & \multicolumn{2}{c}{\textbf{\#Comparisons}} && 
        \multirow{ 2}{*}{\textbf{Use Case}}\\\cmidrule{4-5}
        
        & & & Mated & Non-mated && \\ 
        \midrule
        Adience~\cite{adience} & $19{\small,}370$ & $2{\small,}284$ & $20{\small,}000$ & $20{\small,}000$ && \multirow{2}{*}{General}   \\
        LFW~\cite{lfw} & $13{\small,}233$ & $5{\small,}749$ & $3{\small,}000$ & $3{\small,}000$ && \\
        \midrule
        CPLFW~\cite{cplfw} & $11{\small,}652$ & $3{\small,}930$& $3{\small,}000$ & $3{\small,}000$ && \multirow{2}{*}{Cross-Pose}  \\
        CFP-FP~\cite{cfp-fp} & $7{\small,}000$ & $500$ & $3{\small,}500$ & $3{\small,}500$ && \\
        \midrule
        CALFW~\cite{calfw} & $12{\small,}174$ & $4{\small,}025$ & $3{\small,}000$ & $3{\small,}000$ && \multirow{1}{*}{Cross-Age} \\
        \midrule
        XQLFW~\cite{xqlfw} & $13{\small,}233$ & $5{\small,}749$ & $3{\small,}000$ & $3{\small,}000$ && \multirow{1}{*}{Cross-Quality}\\
        \bottomrule
    \end{tabular}
    }
    \label{tab:my_flips}\vspace{-2mm}
\end{table}

\begin{table}[!t!]
    \centering
    \caption{{Experiments using properly aligned samples.} 
    }
    \renewcommand{\arraystretch}{1.01}
    \resizebox{\columnwidth}{!}{%
\begin{NiceTabularX}{\columnwidth}{c l l | c c c c c c | r}

\toprule
{\Block[fill=brown!10,l]{24-1}{\rotate \textbf{Cross-Model}}}
 & \Block{1-9} {\textbf{AdaFace - pAUC@FMR=}$\mathbf{10^{-3}} (\downarrow)$} \\
\cmidrule{2-10}
 &  &  & \textbf{Adience} & \textbf{LFW} & \textbf{CPLFW} & \textbf{CFP-FP} & \textbf{CALFW} & \textbf{XQLFW} & $\overline{pAUC}$ \\ 
\cmidrule{2-10}
 & \multirow{2}{*}{\textbf{SER-FIQ}~\cite{serfiq}} & B &  $0.839$ &  $0.897*$ &  $0.743*$ &  $0.619*$ &  $0.879*$ &  $0.843$ &  $0.803$  \\
 &  & T &  $\mathbf{0.801}*$ &  $0.915$ &  $0.755$ &  $0.620$ &  $0.915$ &  $0.686*$ & \cellcolor{green!10} $0.782$  \\
\cmidrule{2-10}
 & \multirow{2}{*}{\textbf{DifFIQA(R)}~\cite{diffiqar}} & B &  $0.816*$ &  $0.848$ &  $0.680*$ &  $0.540*$ &  $0.919$ &  $\mathbf{0.610}*$ &  $0.735$  \\
 &  & T &  $0.835$ &  $\mathbf{0.746}*$ &  $0.692$ &  $0.551$ &  $0.882*$ &  $0.643$ & \cellcolor{green!10} $0.725$  \\
\cmidrule{2-10}
 & \multirow{2}{*}{\textbf{CR-FIQA}~\cite{crfiqa}} & B &  $0.844$ &  $0.851$ &  $0.671$ &  $0.544$ &  $\mathbf{0.856}*$ &  $0.685$ &  $0.742$  \\
 &  & T &  $0.818*$ &  $0.832*$ &  $\mathbf{0.664}*$ &  $\mathbf{0.520}*$ &  $0.891$ &  $0.641*$ & \cellcolor{green!10} $0.728$  \\
\cmidrule{2-10}

 & \Block{1-9} {\textbf{SwinFace - pAUC@FMR=}$\mathbf{10^{-3}} (\downarrow)$} \\
\cmidrule{2-10}
 &  &  & \textbf{Adience} & \textbf{LFW} & \textbf{CPLFW} & \textbf{CFP-FP} & \textbf{CALFW} & \textbf{XQLFW} & $\overline{pAUC}$ \\ 
\cmidrule{2-10}
 & \multirow{2}{*}{\textbf{SER-FIQ}~\cite{serfiq}} & B &  $0.811$ &  $0.887*$ &  $\mathbf{0.698}*$ &  $0.534*$ &  $0.867$ &  $0.872$ &  $0.778$  \\
 &  & T &  $\mathbf{0.761}*$ &  $0.939$ &  $0.759$ &  $0.546$ &  $0.840*$ &  $0.636*$ & \cellcolor{green!10} $0.747$  \\
\cmidrule{2-10}
 & \multirow{2}{*}{\textbf{DifFIQA(R)}~\cite{diffiqar}} & B &  $0.805*$ &  $0.859$ &  $0.736$ &  $0.477$ &  $0.897$ &  $\mathbf{0.567}*$ &  $0.724$  \\
 &  & T &  $0.811$ &  $\mathbf{0.770}*$ &  $0.734*$ &  $0.459*$ &  $0.851*$ &  $0.633$ & \cellcolor{green!10} $0.710$  \\
\cmidrule{2-10}
 & \multirow{2}{*}{\textbf{CR-FIQA}~\cite{crfiqa}} & B &  $0.807$ &  $0.879$ &  $0.724$ &  $0.422$ &  $\mathbf{0.813}*$ &  $0.640$ &  $0.714$  \\
 &  & T &  $0.788*$ &  $0.860*$ &  $0.718*$ &  $\mathbf{0.399}*$ &  $0.890$ &  $0.621*$ & \cellcolor{green!10} $0.713$  \\
\cmidrule{2-10}

 & \Block{1-9} {\textbf{TransFace - pAUC@FMR=}$\mathbf{10^{-3}} (\downarrow)$} \\
\cmidrule{2-10}
 &  &  & \textbf{Adience} & \textbf{LFW} & \textbf{CPLFW} & \textbf{CFP-FP} & \textbf{CALFW} & \textbf{XQLFW} & $\overline{pAUC}$ \\ 
\cmidrule{2-10}
 & \multirow{2}{*}{\textbf{SER-FIQ}~\cite{serfiq}} & B &  $0.837$ &  $0.897*$ &  $0.730$ &  $0.657$ &  $0.910*$ &  $0.820$ &  $0.808$  \\
 &  & T &  $\mathbf{0.771}*$ &  $0.915$ &  $0.721*$ &  $0.630*$ &  $0.920$ &  $0.611*$ & \cellcolor{green!10} $0.761$  \\
\cmidrule{2-10}
 & \multirow{2}{*}{\textbf{DifFIQA(R)}~\cite{diffiqar}} & B &  $0.812*$ &  $0.870$ &  $0.640*$ &  $0.528*$ &  $0.920$ &  $\mathbf{0.524}*$ &  $0.716$  \\
 &  & T &  $0.835$ &  $\mathbf{0.784}*$ &  $0.647$ &  $0.551$ &  $\mathbf{0.887}*$ &  $0.566$ & \cellcolor{green!10} $0.712$  \\
\cmidrule{2-10}
 & \multirow{2}{*}{\textbf{CR-FIQA}~\cite{crfiqa}} & B &  $0.829$ &  $0.851$ &  $0.639$ &  $0.512$ &  $0.887*$ &  $0.580*$ &  $0.716$  \\
 &  & T &  $0.807*$ &  $0.840*$ &  $\mathbf{0.625}*$ &  $\mathbf{0.486}*$ &  $0.935$ &  $0.602$ & \cellcolor{green!10} $0.716$  \\

\bottomrule
\toprule
 {\Block[fill=magenta!10,l]{8-1}{\rotate \textbf{Same-Model}}} & \Block{1-9} {\textbf{CosFace - pAUC@FMR=}$\mathbf{10^{-3}} (\downarrow)$} \\
\cmidrule{2-10}
 &  &  & \textbf{Adience} & \textbf{LFW} & \textbf{CPLFW} & \textbf{CFP-FP} & \textbf{CALFW} & \textbf{XQLFW} & $\overline{pAUC}$ \\ 
\cmidrule{2-10}
 & \multirow{2}{*}{\textbf{SER-FIQ}~\cite{serfiq}} & B &  $0.825$ &  $0.858*$ &  $0.760*$ &  $0.606*$ &  $0.908*$ &  $0.770$ &  $0.788$  \\
 &  & T &  $\mathbf{0.791}*$ &  $0.938$ &  $0.789$ &  $0.614$ &  $0.922$ &  $0.563*$ & \cellcolor{green!10} $0.769$  \\
\cmidrule{2-10}
 & \multirow{2}{*}{\textbf{DifFIQA(R)}~\cite{diffiqar}} & B &  $0.805*$ &  $0.831$ &  $0.707*$ &  $0.522$ &  $0.916$ &  $\mathbf{0.557}*$ &  $0.723$  \\
 &  & T &  $0.807$ &  $\mathbf{0.746}*$ &  $0.710$ &  $0.517*$ &  $0.893*$ &  $0.566$ & \cellcolor{green!10} $0.707$  \\
\cmidrule{2-10}
 & \multirow{2}{*}{\textbf{CR-FIQA}~\cite{crfiqa}} & B &  $0.835$ &  $0.851$ &  $0.696$ &  $0.503$ &  $\mathbf{0.889}*$ &  $0.631$ &  $0.734$  \\
 &  & T &  $0.803*$ &  $0.847*$ &  $\mathbf{0.693}*$ &  $\mathbf{0.477}*$ &  $0.934$ &  $0.580*$ & \cellcolor{green!10} $0.722$  \\

\bottomrule

\multicolumn{10}{l}{\textbf{B} \textit{- performance of the baseline approach}, \textbf{T} - \textit{performance of the extended AI-KD approach}}
\end{NiceTabularX}
}
        \label{tab:pauc_analysis_proper_align}\vspace{-3mm}
\end{table}

\section{Experiments \& Results}\label{sec:experiments_and_results}

\noindent\textbf{Experimental setting.} We analyze the performance of AI-KD over 3 FIQA methods and in comparison to $7$ state-of-the-art competitors: $(i)$  the \textbf{unsupervised} FaceQAN~\cite{faceqan} and SER-FIQ~\cite{serfiq} models, $(ii)$ the \textbf{supervised}  FaceQnet~\cite{faceqnet1}, SDD-FIQA \cite{sddfiqa} and DifFIQA(R)~\cite{diffiqar} techniques, and $(iii)$ the \textbf{quality-aware}  MagFace~\cite{magface} and CR-FIQA \cite{crfiqa} methods. We test all methods on $6$ commonly used benchmarks with different quality characteristics, as summarized in Table~\ref{tab:my_flips}, i.e.: Adience~\cite{adience}, Labeled Faces in the Wild (LFW) \cite{lfw}, Cross-Pose Labeled Faces in the Wild (CPLFW) \cite{cplfw}, Celebrities in Frontal-Profile in the Wild (CFP-FP) \cite{cfp-fp}, Cross-Age Labeled Faces in the Wild (CALFW) \cite{calfw} and the Cross-Quality Labeled Faces in the Wild~(XQLFW) \cite{xqlfw}. Because the performance of FIQA techniques is dependent on the FR model used, we investigate how well the techniques generalize over $4$ state-of-the-art models divided into CNN-based models, i.e., AdaFace\footnote{\label{fnote:adaface}\scriptsize{\url{https://github.com/mk-minchul/AdaFace}; \url{https://github.com/deepinsight/insightface}; \url{https://github.com/lxq1000/SwinFace}; \url{https://github.com/DanJun6737/TransFace}}} \cite{adaface}, and CosFace\footref{fnote:adaface}
and Transformer-based models i.e., SwinFace\footref{fnote:adaface} \cite{swinface}, and TransFace\footref{fnote:adaface} \cite{transface}. To evaluate the effects of alignment on the performance of FIQA techniques, we employ four different face landmark detection methods i.e., RetinaFace~\cite{retinaface} (using ResNet50 and {\color{violet}MobileNet} backbones), {\color{olive}MTCNN}~\cite{mtcnn}, and {\color{teal}DLib}~\cite{dlib}.

\begin{table}[!t!]
    \centering
    \caption{{Same-Model Experiments using misaligned samples.} 
    }
    \renewcommand{\arraystretch}{1.01}
    \resizebox{\columnwidth}{!}{%
\begin{NiceTabularX}{\columnwidth}{l l l | c c c c c c | r}

\toprule
& \Block{1-9} {\textbf{CosFace - pAUC@FMR=}$\mathbf{10^{-3}} (\downarrow)$} \\
\cmidrule{1-10}
{\Block[fill=violet!10,l]{7-1}{\rotate \textbf{RetinaFace(MNet)}}}  &  &  & \textbf{Adience} & \textbf{LFW} & \textbf{CPLFW} & \textbf{CFP-FP} & \textbf{CALFW} & \textbf{XQLFW} & $\overline{pAUC}$ \\ 
\cmidrule{2-10}
 & \multirow{2}{*}{\textbf{SER-FIQ}~\cite{serfiq}} & B &  $0.868$ &  $\mathbf{0.813}*$ &  $0.747*$ &  $0.614$ &  $0.896*$ &  $0.744$ &  $0.780$  \\
 &  & T &  $\mathbf{0.832}*$ &  $0.938$ &  $0.792$ &  $0.586*$ &  $0.920$ &  $0.583*$ & \cellcolor{green!10} $0.775$  \\
\cmidrule{2-10}
 & \multirow{2}{*}{\textbf{DifFIQA(R)}~\cite{diffiqar}} & B &  $0.861$ &  $0.921$ &  $0.699*$ &  $0.515$ &  $\mathbf{0.894}*$ &  $\mathbf{0.551}*$ &  $0.740$  \\
 &  & T &  $0.840*$ &  $0.825*$ &  $0.709$ &  $0.499*$ &  $0.901$ &  $0.562$ & \cellcolor{green!10} $0.723$  \\
\cmidrule{2-10}
 & \multirow{2}{*}{\textbf{CR-FIQA}~\cite{crfiqa}} & B &  $0.865$ &  $0.851*$ &  $\mathbf{0.689}*$ &  $0.508$ &  $0.895*$ &  $0.638$ &  $0.741$  \\
 &  & T &  $0.836*$ &  $0.870$ &  $0.699$ &  $\mathbf{0.472}*$ &  $0.931$ &  $0.585*$ & \cellcolor{green!10} $0.732$  \\

\toprule
 {\Block[fill=olive!10,l]{7-1}{\rotate \textbf{MTCNN}}} &  &  & \textbf{Adience} & \textbf{LFW} & \textbf{CPLFW} & \textbf{CFP-FP} & \textbf{CALFW} & \textbf{XQLFW} & $\overline{pAUC}$ \\ 
\cmidrule{2-10}
 & \multirow{2}{*}{\textbf{SER-FIQ}~\cite{serfiq}} & B &  $0.894*$ &  $0.871*$ &  $0.837$ &  $0.734$ &  $0.909$ &  $0.692$ &  $0.823$  \\
 &  & T &  $0.899$ &  $0.886$ &  $0.773*$ &  $0.695*$ &  $0.899*$ &  $0.590*$ & \cellcolor{green!10} $0.790$  \\
\cmidrule{2-10}
 & \multirow{2}{*}{\textbf{DifFIQA(R)}~\cite{diffiqar}} & B &  $0.889*$ &  $0.799$ &  $0.727$ &  $0.713$ &  $0.912$ &  $\mathbf{0.550}*$ &  $0.765$  \\
 &  & T &  $0.906$ &  $\mathbf{0.700}*$ &  $0.715*$ &  $0.692*$ &  $0.905*$ &  $0.564$ & \cellcolor{green!10} $0.747$  \\
\cmidrule{2-10}
 & \multirow{2}{*}{\textbf{CR-FIQA}~\cite{crfiqa}} & B &  $0.915$ &  $0.808$ &  $0.718$ &  $0.662$ &  $\mathbf{0.889}*$ &  $0.605$ &  $0.766$  \\
 &  & T &  $\mathbf{0.873}*$ &  $0.743*$ &  $\mathbf{0.709}*$ &  $\mathbf{0.622}*$ &  $0.934$ &  $0.601*$ & \cellcolor{green!10} $0.747$  \\

\toprule
{\Block[fill=teal!10,l]{7-1}{\rotate \textbf{DLib}}} &  &  & \textbf{Adience} & \textbf{LFW} & \textbf{CPLFW} & \textbf{CFP-FP} & \textbf{CALFW} & \textbf{XQLFW} & $\overline{pAUC}$ \\ 
\cmidrule{2-10}
 & \multirow{2}{*}{\textbf{SER-FIQ}~\cite{serfiq}} & B &  $0.910$ &  $0.853$ &  $0.881*$ &  $0.747*$ &  $0.902*$ &  $0.744$ & \cellcolor{green!10} $0.839$  \\
 &  & T &  $0.883*$ &  $0.848*$ &  $0.901$ &  $0.868$ &  $0.907$ &  $0.638*$ &  $0.841$  \\
\cmidrule{2-10}
 & \multirow{2}{*}{\textbf{DifFIQA(R)}~\cite{diffiqar}} & B &  $0.880*$ &  $0.809$ &  $0.939$ &  $0.970$ &  $0.892$ &  $\mathbf{0.615}*$ &  $0.851$  \\
 &  & T &  $0.884$ &  $\mathbf{0.687}*$ &  $\mathbf{0.842}*$ &  $\mathbf{0.722}*$ &  $\mathbf{0.879}*$ &  $0.631$ & \cellcolor{green!10} $0.774$  \\
\cmidrule{2-10}
 & \multirow{2}{*}{\textbf{CR-FIQA}~\cite{crfiqa}} & B &  $0.915$ &  $0.789$ &  $0.902$ &  $0.804$ &  $0.882*$ &  $0.667*$ &  $0.827$  \\
 &  & T &  $\mathbf{0.876}*$ &  $0.735*$ &  $0.864*$ &  $0.755*$ &  $0.910$ &  $0.693$ & \cellcolor{green!10} $0.805$  \\

\bottomrule
\multicolumn{10}{l}{\textbf{B} \textit{- performance of the baseline approach}, \textbf{T} - \textit{performance of the extended AI-KD approach}}

\end{NiceTabularX}
}
        \label{tab:pauc_analysis_misalignment_same_model}\vspace{-3mm}
\end{table}


\noindent\textbf{Evaluation methodology.} Using standard evaluation methodology \cite{faceqan,serfiq,crfiqa}, we quantify the performance of the tested methods using the pAUC~(partial Area Under the Curve) of the Error-versus-Discard Characteristic (EDC) curves (also referred to as Error-versus-Reject Characteristic (ERC) curves). The EDC curves measure how the performance of a given FR model improves, when rejecting some percentage of the lowest quality images from the dataset, and are calculated using a predefined False Match Rate~(FMR) ($10^{-3}$ in our case), while increasing low-quality image discard (reject) rates. In real-world situation, it is not feasible to reject a large percentage of all samples, therefore we report the pAUC at lower values of the discard (reject) rates ($30\%$ in our case). Furthermore, for easier interpretation and comparison of scores over different dataset, we normalize the calculated pAUC values using the FNMR at the $0\%$ discard rate
, with lower pAUC values indicating better performance.

\vspace{1mm}\noindent\textbf{Implementation Details.} We use the VGGFace2 dataset, to train the evaluted models. For the FR backbone of the quality-regression models, we use a ResNet100 model, trained using the CosFace loss. Based on this choice, we split the experiments into Cross- and Same-model scenarios based on whether the evaluated model is also trained using the CosFace loss. For the sample transformation, we use $p=3$, as preliminary testing showed that different landmark detection methods vary between $3$-$4$ pixels in their predictions. To train the model, we used Stochastic Gradient Descend, with a learning rate of $0.05$. Additionally to further improve the model we employed Stochastic Weight Averaging. All experiments were conducted on a desktop PC with an Intel i9-10900KF CPU, $64$ GB of RAM and an Nvidia $3090$ GPU.

\begin{table}[!t!]
    \centering
    \caption{{Cross-Model Experiments using misaligned samples.\vspace{-0.5mm}} 
    }
    \renewcommand{\arraystretch}{1.02}
    \resizebox{\columnwidth}{!}{%
\begin{NiceTabularX}{\columnwidth}{l  l l | c c c c c c | r}

\toprule
{\Block[fill=violet!10,l]{24-1}{\rotate \textbf{RetinaFace(MNet)}}}
 & \Block{1-9} {\textbf{AdaFace - pAUC@FMR=}$\mathbf{10^{-3}} (\downarrow)$} \\
\cmidrule{2-10}
 &  &  & \textbf{Adience} & \textbf{LFW} & \textbf{CPLFW} & \textbf{CFP-FP} & \textbf{CALFW} & \textbf{XQLFW} & $\overline{pAUC}$ \\ 
\cmidrule{2-10}
 & \multirow{2}{*}{\textbf{SER-FIQ}~\cite{serfiq}} & B &  $0.886$ &  $0.852*$ &  $0.738*$ &  $0.634$ &  $0.865*$ &  $0.829$ &  $0.801$  \\
 &  & T &  $\mathbf{0.841}*$ &  $0.923$ &  $0.755$ &  $0.594*$ &  $0.916$ &  $0.694*$ & \cellcolor{green!10} $0.787$  \\
\cmidrule{2-10}
 & \multirow{2}{*}{\textbf{DifFIQA(R)}~\cite{diffiqar}} & B &  $0.880$ &  $0.953$ &  $0.670*$ &  $0.506*$ &  $0.898$ &  $\mathbf{0.601}*$ &  $0.751$  \\
 &  & T &  $0.866*$ &  $\mathbf{0.825}*$ &  $0.691$ &  $0.507$ &  $0.884*$ &  $0.635$ & \cellcolor{green!10} $0.735$  \\
\cmidrule{2-10}
 & \multirow{2}{*}{\textbf{CR-FIQA}~\cite{crfiqa}} & B &  $0.877$ &  $0.844*$ &  $\mathbf{0.659}*$ &  $0.540$ &  $\mathbf{0.863}*$ &  $0.683$ &  $0.744$  \\
 &  & T &  $0.847*$ &  $0.855$ &  $0.665$ &  $\mathbf{0.501}*$ &  $0.893$ &  $0.649*$ & \cellcolor{green!10} $0.735$  \\
\cmidrule{2-10}

 & \Block{1-9} {\textbf{SwinFace - pAUC@FMR=}$\mathbf{10^{-3}} (\downarrow)$} \\
\cmidrule{2-10}
 &  &  & \textbf{Adience} & \textbf{LFW} & \textbf{CPLFW} & \textbf{CFP-FP} & \textbf{CALFW} & \textbf{XQLFW} & $\overline{pAUC}$ \\ 
\cmidrule{2-10}
 & \multirow{2}{*}{\textbf{SER-FIQ}~\cite{serfiq}} & B &  $0.854$ &  $\mathbf{0.840}*$ &  $0.712*$ &  $0.500*$ &  $0.854$ &  $0.794$ &  $0.759$  \\
 &  & T &  $\mathbf{0.810}*$ &  $0.947$ &  $0.778$ &  $0.540$ &  $0.837*$ &  $0.641*$ & \cellcolor{green!10} $0.759$  \\
\cmidrule{2-10}
 & \multirow{2}{*}{\textbf{DifFIQA(R)}~\cite{diffiqar}} & B &  $0.852$ &  $0.945$ &  $0.728*$ &  $0.441*$ &  $0.842*$ &  $\mathbf{0.586}*$ & \cellcolor{green!10} $0.732$  \\
 &  & T &  $0.836*$ &  $0.852*$ &  $0.736$ &  $0.441$ &  $0.870$ &  $0.686$ &  $0.737$  \\
\cmidrule{2-10}
 & \multirow{2}{*}{\textbf{CR-FIQA}~\cite{crfiqa}} & B &  $0.841$ &  $0.872*$ &  $\mathbf{0.702}*$ &  $0.414$ &  $\mathbf{0.816}*$ &  $0.672$ & \cellcolor{green!10} $0.720$  \\
 &  & T &  $0.820*$ &  $0.877$ &  $0.725$ &  $\mathbf{0.400}*$ &  $0.891$ &  $0.620*$ &  $0.722$  \\
\cmidrule{2-10}

 & \Block{1-9} {\textbf{TransFace - pAUC@FMR=}$\mathbf{10^{-3}} (\downarrow)$} \\
\cmidrule{2-10}
 &  &  & \textbf{Adience} & \textbf{LFW} & \textbf{CPLFW} & \textbf{CFP-FP} & \textbf{CALFW} & \textbf{XQLFW} & $\overline{pAUC}$ \\ 
\cmidrule{2-10}
 & \multirow{2}{*}{\textbf{SER-FIQ}~\cite{serfiq}} & B &  $0.886$ &  $0.852*$ &  $0.715*$ &  $0.641$ &  $0.901*$ &  $0.801$ &  $0.799$  \\
 &  & T &  $\mathbf{0.818}*$ &  $0.923$ &  $0.725$ &  $0.605*$ &  $0.919$ &  $0.609*$ & \cellcolor{green!10} $0.766$  \\
\cmidrule{2-10}
 & \multirow{2}{*}{\textbf{DifFIQA(R)}~\cite{diffiqar}} & B &  $0.882$ &  $0.953$ &  $0.632*$ &  $0.521$ &  $0.898$ &  $\mathbf{0.534}*$ &  $0.737$  \\
 &  & T &  $0.859*$ &  $0.847*$ &  $0.647$ &  $0.510*$ &  $\mathbf{0.890}*$ &  $0.551$ & \cellcolor{green!10} $0.717$  \\
\cmidrule{2-10}
 & \multirow{2}{*}{\textbf{CR-FIQA}~\cite{crfiqa}} & B &  $0.864$ &  $\mathbf{0.844}*$ &  $0.634$ &  $0.521$ &  $0.896*$ &  $0.588*$ &  $0.724$  \\
 &  & T &  $0.843*$ &  $0.855$ &  $\mathbf{0.628}*$ &  $\mathbf{0.469}*$ &  $0.933$ &  $0.603$ & \cellcolor{green!10} $0.722$  \\

\bottomrule
\toprule
{\Block[fill=olive!10,l]{24-1}{\rotate \textbf{MTCNN}}}
 & \Block{1-9} {\textbf{AdaFace - pAUC@FMR=}$\mathbf{10^{-3}} (\downarrow)$} \\
\cmidrule{2-10}
 &  &  & \textbf{Adience} & \textbf{LFW} & \textbf{CPLFW} & \textbf{CFP-FP} & \textbf{CALFW} & \textbf{XQLFW} & $\overline{pAUC}$ \\ 
\cmidrule{2-10}
 & \multirow{2}{*}{\textbf{SER-FIQ}~\cite{serfiq}} & B &  $0.912$ &  $0.910*$ &  $0.802$ &  $0.708*$ &  $0.878*$ &  $0.846$ &  $0.843$  \\
 &  & T &  $0.909*$ &  $0.911$ &  $0.739*$ &  $0.742$ &  $0.890$ &  $0.697*$ & \cellcolor{green!10} $0.815$  \\
\cmidrule{2-10}
 & \multirow{2}{*}{\textbf{DifFIQA(R)}~\cite{diffiqar}} & B &  $0.917*$ &  $0.815$ &  $0.695$ &  $0.745$ &  $0.910$ &  $\mathbf{0.622}*$ &  $0.784$  \\
 &  & T &  $0.933$ &  $\mathbf{0.683}*$ &  $0.686*$ &  $0.729*$ &  $0.895*$ &  $0.655$ & \cellcolor{green!10} $0.763$  \\
\cmidrule{2-10}
 & \multirow{2}{*}{\textbf{CR-FIQA}~\cite{crfiqa}} & B &  $0.929$ &  $0.832$ &  $0.685$ &  $0.713$ &  $\mathbf{0.857}*$ &  $0.661$ &  $0.779$  \\
 &  & T &  $\mathbf{0.891}*$ &  $0.768*$ &  $\mathbf{0.678}*$ &  $\mathbf{0.658}*$ &  $0.895$ &  $0.650*$ & \cellcolor{green!10} $0.757$  \\
\cmidrule{2-10}

 & \Block{1-9} {\textbf{SwinFace - pAUC@FMR=}$\mathbf{10^{-3}} (\downarrow)$} \\
\cmidrule{2-10}
 &  &  & \textbf{Adience} & \textbf{LFW} & \textbf{CPLFW} & \textbf{CFP-FP} & \textbf{CALFW} & \textbf{XQLFW} & $\overline{pAUC}$ \\ 
\cmidrule{2-10}
 & \multirow{2}{*}{\textbf{SER-FIQ}~\cite{serfiq}} & B &  $0.883$ &  $0.901*$ &  $0.833$ &  $0.632$ &  $0.872$ &  $0.832$ &  $0.826$  \\
 &  & T &  $0.875*$ &  $0.901$ &  $0.777*$ &  $0.601*$ &  $0.823*$ &  $0.681*$ & \cellcolor{green!10} $0.776$  \\
\cmidrule{2-10}
 & \multirow{2}{*}{\textbf{DifFIQA(R)}~\cite{diffiqar}} & B &  $0.892*$ &  $0.825$ &  $0.745$ &  $0.613$ &  $0.915$ &  $0.564$ &  $0.759$  \\
 &  & T &  $0.903$ &  $\mathbf{0.723}*$ &  $0.732*$ &  $0.545*$ &  $0.896*$ &  $\mathbf{0.563}*$ & \cellcolor{green!10} $0.727$  \\
\cmidrule{2-10}
 & \multirow{2}{*}{\textbf{CR-FIQA}~\cite{crfiqa}} & B &  $0.894$ &  $0.827$ &  $0.732$ &  $0.518$ &  $\mathbf{0.813}*$ &  $0.605*$ &  $0.732$  \\
 &  & T &  $\mathbf{0.862}*$ &  $0.752*$ &  $\mathbf{0.725}*$ &  $\mathbf{0.482}*$ &  $0.892$ &  $0.621$ & \cellcolor{green!10} $0.722$  \\
\cmidrule{2-10}

 & \Block{1-9} {\textbf{TransFace - pAUC@FMR=}$\mathbf{10^{-3}} (\downarrow)$} \\
\cmidrule{2-10}
 &  &  & \textbf{Adience} & \textbf{LFW} & \textbf{CPLFW} & \textbf{CFP-FP} & \textbf{CALFW} & \textbf{XQLFW} & $\overline{pAUC}$ \\ 
\cmidrule{2-10}
 & \multirow{2}{*}{\textbf{SER-FIQ}~\cite{serfiq}} & B &  $0.906$ &  $0.910*$ &  $0.801$ &  $0.748$ &  $0.910$ &  $0.794$ &  $0.845$  \\
 &  & T &  $\mathbf{0.875}*$ &  $0.911$ &  $0.714*$ &  $0.682*$ &  $0.900*$ &  $0.619*$ & \cellcolor{green!10} $0.783$  \\
\cmidrule{2-10}
 & \multirow{2}{*}{\textbf{DifFIQA(R)}~\cite{diffiqar}} & B &  $0.904*$ &  $0.837$ &  $0.672$ &  $0.712$ &  $0.922$ &  $\mathbf{0.540}*$ &  $0.764$  \\
 &  & T &  $0.925$ &  $\mathbf{0.722}*$ &  $0.658*$ &  $0.685*$ &  $0.897*$ &  $0.545$ & \cellcolor{green!10} $0.739$  \\
\cmidrule{2-10}
 & \multirow{2}{*}{\textbf{CR-FIQA}~\cite{crfiqa}} & B &  $0.917$ &  $0.840$ &  $0.659$ &  $0.657$ &  $\mathbf{0.888}*$ &  $0.576*$ &  $0.756$  \\
 &  & T &  $0.878*$ &  $0.768*$ &  $\mathbf{0.645}*$ &  $\mathbf{0.589}*$ &  $0.932$ &  $0.594$ & \cellcolor{green!10} $0.734$  \\

\bottomrule
\toprule
{\Block[fill=teal!10,l]{24-1}{\rotate \textbf{DLib}}}
 & \Block{1-9} {\textbf{AdaFace - pAUC@FMR=}$\mathbf{10^{-3}} (\downarrow)$} \\
\cmidrule{2-10}
 &  &  & \textbf{Adience} & \textbf{LFW} & \textbf{CPLFW} & \textbf{CFP-FP} & \textbf{CALFW} & \textbf{XQLFW} & $\overline{pAUC}$ \\ 
\cmidrule{2-10}
 & \multirow{2}{*}{\textbf{SER-FIQ}~\cite{serfiq}} & B &  $0.928$ &  $0.885$ &  $0.858*$ &  $0.736*$ &  $0.884*$ &  $0.868$ &  $0.860$  \\
 &  & T &  $0.901*$ &  $0.848*$ &  $0.906$ &  $0.769$ &  $0.895$ &  $0.765*$ & \cellcolor{green!10} $0.847$  \\
\cmidrule{2-10}
 & \multirow{2}{*}{\textbf{DifFIQA(R)}~\cite{diffiqar}} & B &  $0.902*$ &  $0.841$ &  $0.955$ &  $0.871$ &  $0.888$ &  $\mathbf{0.718}*$ &  $0.863$  \\
 &  & T &  $0.906$ &  $\mathbf{0.696}*$ &  $\mathbf{0.832}*$ &  $\mathbf{0.699}*$ &  $0.875*$ &  $0.724$ & \cellcolor{green!10} $0.789$  \\
\cmidrule{2-10}
 & \multirow{2}{*}{\textbf{CR-FIQA}~\cite{crfiqa}} & B &  $0.930$ &  $0.821$ &  $0.881$ &  $0.776$ &  $\mathbf{0.854}*$ &  $0.748*$ &  $0.835$  \\
 &  & T &  $\mathbf{0.897}*$ &  $0.752*$ &  $0.860*$ &  $0.728*$ &  $0.875$ &  $0.751$ & \cellcolor{green!10} $0.810$  \\
\cmidrule{2-10}

 & \Block{1-9} {\textbf{SwinFace - pAUC@FMR=}$\mathbf{10^{-3}} (\downarrow)$} \\
\cmidrule{2-10}
 &  &  & \textbf{Adience} & \textbf{LFW} & \textbf{CPLFW} & \textbf{CFP-FP} & \textbf{CALFW} & \textbf{XQLFW} & $\overline{pAUC}$ \\ 
\cmidrule{2-10}
 & \multirow{2}{*}{\textbf{SER-FIQ}~\cite{serfiq}} & B &  $0.908$ &  $0.874$ &  $\mathbf{0.801}*$ &  $0.717*$ &  $0.868$ &  $0.727$ &  $0.816$  \\
 &  & T &  $\mathbf{0.861}*$ &  $0.836*$ &  $0.869$ &  $0.727$ &  $0.844*$ &  $0.652*$ & \cellcolor{green!10} $0.798$  \\
\cmidrule{2-10}
 & \multirow{2}{*}{\textbf{DifFIQA(R)}~\cite{diffiqar}} & B &  $0.873*$ &  $0.836$ &  $0.847$ &  $0.875$ &  $0.875$ &  $\mathbf{0.624}*$ &  $0.822$  \\
 &  & T &  $0.873$ &  $\mathbf{0.709}*$ &  $0.807*$ &  $\mathbf{0.631}*$ &  $0.843*$ &  $0.641$ & \cellcolor{green!10} $0.751$  \\
\cmidrule{2-10}
 & \multirow{2}{*}{\textbf{CR-FIQA}~\cite{crfiqa}} & B &  $0.891$ &  $0.808$ &  $0.837$ &  $0.699$ &  $\mathbf{0.828}*$ &  $0.707$ &  $0.795$  \\
 &  & T &  $0.866*$ &  $0.735*$ &  $0.825*$ &  $0.638*$ &  $0.877$ &  $0.678*$ & \cellcolor{green!10} $0.770$  \\
\cmidrule{2-10}

 & \Block{1-9} {\textbf{TransFace - pAUC@FMR=}$\mathbf{10^{-3}} (\downarrow)$} \\
\cmidrule{2-10}
 &  &  & \textbf{Adience} & \textbf{LFW} & \textbf{CPLFW} & \textbf{CFP-FP} & \textbf{CALFW} & \textbf{XQLFW} & $\overline{pAUC}$ \\ 
\cmidrule{2-10}
 & \multirow{2}{*}{\textbf{SER-FIQ}~\cite{serfiq}} & B &  $0.919$ &  $0.885$ &  $0.859*$ &  $0.763*$ &  $0.908$ &  $0.806$ &  $0.857$  \\
 &  & T &  $\mathbf{0.870}*$ &  $0.848*$ &  $0.899$ &  $0.804$ &  $0.906*$ &  $0.688*$ & \cellcolor{green!10} $0.836$  \\
\cmidrule{2-10}
 & \multirow{2}{*}{\textbf{DifFIQA(R)}~\cite{diffiqar}} & B &  $0.882*$ &  $0.848$ &  $0.919$ &  $0.944$ &  $0.900$ &  $0.620$ &  $0.852$  \\
 &  & T &  $0.887$ &  $\mathbf{0.726}*$ &  $\mathbf{0.801}*$ &  $\mathbf{0.685}*$ &  $\mathbf{0.872}*$ &  $\mathbf{0.616}*$ & \cellcolor{green!10} $0.765$  \\
\cmidrule{2-10}
 & \multirow{2}{*}{\textbf{CR-FIQA}~\cite{crfiqa}} & B &  $0.901$ &  $0.821$ &  $0.872$ &  $0.840$ &  $0.879*$ &  $0.647*$ &  $0.827$  \\
 &  & T &  $0.881*$ &  $0.752*$ &  $0.834*$ &  $0.745*$ &  $0.913$ &  $0.657$ & \cellcolor{green!10} $0.797$  \\

\bottomrule

\multicolumn{10}{l}{\textbf{B} \textit{- performance of the baseline approach}, \textbf{T} - \textit{performance of the extended AI-KD approach}}
\end{NiceTabularX}
}
        \label{tab:pauc_analysis_misalignment_cross_model}\vspace{-4mm}
\end{table}

\begin{table}[!t!]
    \centering
    \caption{{Comparison with state-of-the-art -- properly aligned samples.} 
    }
    \renewcommand{\arraystretch}{1.05}
    \resizebox{\columnwidth}{!}{%
\begin{NiceTabularX}{\columnwidth}{l l | c c c c c c | r}

\toprule
{\Block[fill=brown!10,l]{36-1}{\rotate \textbf{Cross-Model}}} & \multicolumn{8}{c}{\textbf{AdaFace}} \\
\cmidrule{2-9}
 & & \textbf{Adience} & \textbf{LFW} & \textbf{CPLFW} & \textbf{CFP-FP} & \textbf{CALFW} & \textbf{XQLFW} & $\overline{pAUC}$ \\ 
\cmidrule{2-9}
 & \textbf{FaceQAN}~\cite{faceqan} &  $0.880$ &  $0.780$ &  $0.679$ &  $\mathbf{0.387}$ &  $0.952$ &  $0.634$ & \cellcolor{green!10} $0.719$  \\
 & \textbf{SER-FIQ}~\cite{serfiq} &  $0.839$ &  $0.897$ &  $0.743$ &  $0.619$ &  $0.879$ &  $0.843$ &  $0.803$  \\
 & \textbf{FaceQnet}~\cite{faceqnet} &  $0.961$ &  $0.862$ &  $0.883$ &  $0.735$ &  $0.937$ &  $0.977$ &  $0.893$  \\
 & \textbf{SDD-FIQA}~\cite{sddfiqa} &  $0.839$ &  $0.897$ &  $0.743$ &  $0.619$ &  $0.879$ &  $0.843$ &  $0.803$  \\
 & \textbf{DifFIQA(R)}~\cite{diffiqar} &  $0.816$ &  $0.848$ &  $0.680$ &  $0.540$ &  $0.919$ &  $\mathbf{0.610}$ &  $0.735$  \\
 & \textbf{MagFace}~\cite{magface} &  $0.862$ &  $0.874$ &  $0.735$ &  $0.692$ &  $0.868$ &  $0.914$ &  $0.824$  \\
 & \textbf{CR-FIQA}~\cite{crfiqa} &  $0.844$ &  $0.851$ &  $0.671$ &  $0.544$ &  $\mathbf{0.856}$ &  $0.685$ &  $0.742$  \\
\cmidrule{2-9}
 & \textbf{AI-KD(SER-FIQ)} &  $\mathbf{0.801}$ &  $0.915$ &  $0.755$ &  $0.620$ &  $0.915$ &  $0.686$ &  $0.782$  \\
 & \textbf{AI-KD(CR-FIQA)} &  $0.818$ &  $0.832$ &  $\mathbf{0.664}$ &  $0.520$ &  $0.891$ &  $0.641$ & \cellcolor{magenta!10} $0.728$  \\
 & \textbf{AI-KD(DifFIQA(R))} &  $0.835$ &  $\mathbf{0.746}$ &  $0.692$ &  $0.551$ &  $0.882$ &  $0.643$ & \cellcolor{cyan!10} $0.725$  \\
\cmidrule{2-9}

 & \multicolumn{8}{c}{\textbf{SwinFace}} \\
\cmidrule{2-9}
 & & \textbf{Adience} & \textbf{LFW} & \textbf{CPLFW} & \textbf{CFP-FP} & \textbf{CALFW} & \textbf{XQLFW} & $\overline{pAUC}$ \\ 
\cmidrule{2-9}
 & \textbf{FaceQAN}~\cite{faceqan} &  $0.860$ &  $0.797$ &  $0.702$ &  $0.409$ &  $0.966$ &  $0.613$ &  $0.725$  \\
 & \textbf{SER-FIQ}~\cite{serfiq} &  $0.811$ &  $0.887$ &  $\mathbf{0.698}$ &  $0.534$ &  $0.867$ &  $0.872$ &  $0.778$  \\
 & \textbf{FaceQnet}~\cite{faceqnet} &  $0.918$ &  $0.891$ &  $0.847$ &  $0.652$ &  $0.938$ &  $0.927$ &  $0.862$  \\
 & \textbf{SDD-FIQA}~\cite{sddfiqa} &  $0.811$ &  $0.887$ &  $0.698$ &  $0.534$ &  $0.867$ &  $0.872$ &  $0.778$  \\
 & \textbf{DifFIQA(R)}~\cite{diffiqar} &  $0.805$ &  $0.859$ &  $0.736$ &  $0.477$ &  $0.897$ &  $\mathbf{0.567}$ &  $0.724$  \\
 & \textbf{MagFace}~\cite{magface} &  $0.830$ &  $0.888$ &  $0.774$ &  $0.551$ &  $0.855$ &  $0.943$ &  $0.807$  \\
 & \textbf{CR-FIQA}~\cite{crfiqa} &  $0.807$ &  $0.879$ &  $0.724$ &  $0.422$ &  $\mathbf{0.813}$ &  $0.640$ & \cellcolor{magenta!10} $0.714$  \\
\cmidrule{2-9}
 & \textbf{AI-KD(SER-FIQ)} &  $\mathbf{0.761}$ &  $0.939$ &  $0.759$ &  $0.546$ &  $0.840$ &  $0.636$ &  $0.747$  \\
 & \textbf{AI-KD(CR-FIQA)} &  $0.788$ &  $0.860$ &  $0.718$ &  $\mathbf{0.399}$ &  $0.890$ &  $0.621$ & \cellcolor{cyan!10} $0.713$  \\
 & \textbf{AI-KD(DifFIQA(R))} &  $0.811$ &  $\mathbf{0.770}$ &  $0.734$ &  $0.459$ &  $0.851$ &  $0.633$ & \cellcolor{green!10} $0.710$  \\
\cmidrule{2-9}

 & \multicolumn{8}{c}{\textbf{TransFace}} \\
\cmidrule{2-9}
 & & \textbf{Adience} & \textbf{LFW} & \textbf{CPLFW} & \textbf{CFP-FP} & \textbf{CALFW} & \textbf{XQLFW} & $\overline{pAUC}$ \\ 
\cmidrule{2-9}
 & \textbf{FaceQAN}~\cite{faceqan} &  $0.874$ &  $0.802$ &  $0.633$ &  $\mathbf{0.388}$ &  $0.986$ &  $0.575$ & \cellcolor{green!10} $0.710$  \\
 & \textbf{SER-FIQ}~\cite{serfiq} &  $0.837$ &  $0.897$ &  $0.730$ &  $0.657$ &  $0.910$ &  $0.820$ &  $0.808$  \\
 & \textbf{FaceQnet}~\cite{faceqnet} &  $0.934$ &  $0.862$ &  $0.885$ &  $0.747$ &  $0.965$ &  $1.007$ &  $0.900$  \\
 & \textbf{SDD-FIQA}~\cite{sddfiqa} &  $0.837$ &  $0.897$ &  $0.730$ &  $0.657$ &  $0.910$ &  $0.820$ &  $0.808$  \\
 & \textbf{DifFIQA(R)}~\cite{diffiqar} &  $0.812$ &  $0.870$ &  $0.640$ &  $0.528$ &  $0.920$ &  $\mathbf{0.524}$ & \cellcolor{magenta!10} $0.716$  \\
 & \textbf{MagFace}~\cite{magface} &  $0.869$ &  $0.841$ &  $0.729$ &  $0.652$ &  $0.901$ &  $0.935$ &  $0.821$  \\
 & \textbf{CR-FIQA}~\cite{crfiqa} &  $0.829$ &  $0.851$ &  $0.639$ &  $0.512$ &  $0.887$ &  $0.580$ &  $0.716$  \\
\cmidrule{2-9}
 & \textbf{AI-KD(SER-FIQ)} &  $\mathbf{0.771}$ &  $0.915$ &  $0.721$ &  $0.630$ &  $0.920$ &  $0.611$ &  $0.761$  \\
 & \textbf{AI-KD(CR-FIQA)} &  $0.807$ &  $0.840$ &  $\mathbf{0.625}$ &  $0.486$ &  $0.935$ &  $0.602$ &  $0.716$  \\
 & \textbf{AI-KD(DifFIQA(R))} &  $0.835$ &  $\mathbf{0.784}$ &  $0.647$ &  $0.551$ &  $\mathbf{0.887}$ &  $0.566$ & \cellcolor{cyan!10} $0.712$  \\

 \bottomrule
 \toprule

{\Block[fill=magenta!10,l]{12-1}{\rotate \textbf{Same-Model}}} & \multicolumn{8}{c}{\textbf{CosFace}} \\
\cmidrule{2-9}
 & & \textbf{Adience} & \textbf{LFW} & \textbf{CPLFW} & \textbf{CFP-FP} & \textbf{CALFW} & \textbf{XQLFW} & $\overline{pAUC}$ \\ 
\cmidrule{2-9}
 & \textbf{FaceQAN}~\cite{faceqan} &  $0.866$ &  $0.772$ &  $0.702$ &  $\mathbf{0.374}$ &  $0.987$ &  $0.574$ & \cellcolor{cyan!10} $0.712$  \\
 & \textbf{SER-FIQ}~\cite{serfiq} &  $0.825$ &  $0.858$ &  $0.760$ &  $0.606$ &  $0.908$ &  $0.770$ &  $0.788$  \\
 & \textbf{FaceQnet}~\cite{faceqnet} &  $0.948$ &  $0.862$ &  $0.867$ &  $0.691$ &  $0.956$ &  $0.894$ &  $0.870$  \\
 & \textbf{SDD-FIQA}~\cite{sddfiqa} &  $0.825$ &  $0.858$ &  $0.760$ &  $0.606$ &  $0.908$ &  $0.770$ &  $0.788$  \\
 & \textbf{DifFIQA(R)}~\cite{diffiqar} &  $0.805$ &  $0.831$ &  $0.707$ &  $0.522$ &  $0.916$ &  $\mathbf{0.557}$ &  $0.723$  \\
 & \textbf{MagFace}~\cite{magface} &  $0.854$ &  $0.866$ &  $0.763$ &  $0.711$ &  $0.902$ &  $0.944$ &  $0.840$  \\
 & \textbf{CR-FIQA}~\cite{crfiqa} &  $0.835$ &  $0.851$ &  $0.696$ &  $0.503$ &  $\mathbf{0.889}$ &  $0.631$ &  $0.734$  \\
\cmidrule{2-9}
 & \textbf{AI-KD(SER-FIQ)} &  $\mathbf{0.791}$ &  $0.938$ &  $0.789$ &  $0.614$ &  $0.922$ &  $0.563$ &  $0.769$  \\
 & \textbf{AI-KD(CR-FIQA)} &  $0.803$ &  $0.847$ &  $\mathbf{0.693}$ &  $0.477$ &  $0.934$ &  $0.580$ & \cellcolor{magenta!10} $0.722$  \\
 & \textbf{AI-KD(DifFIQA(R))} &  $0.807$ &  $\mathbf{0.746}$ &  $0.710$ &  $0.517$ &  $0.893$ &  $0.566$ & \cellcolor{green!10} $0.707$  \\

\bottomrule

    \end{NiceTabularX}
}
        \label{tab:pauc_sota_proper_alignment}\vspace{-2mm}
\end{table}

\subsection{Analysis of AI-KD}\label{sec:experiments_and_results:analysis}

In this section, we analyze how the presented AI-KD technique improves the performance of state-of-the-art FIQA methods, across a variety of benchmark datasets and FR models. We chose three distinct FIQA methods i.e.: the unsupervised SER-FIQ, the supervised DifFIQA(R), and the quality-aware CR-FIQA methods. We separate the experiments by two criteria: $(i)$ based on the used FR model for evaluation, and $(ii)$ based on the alignment of the evaluation benchmarks. Based on the FR model we consider Cross- and Same-Model experiments, where in the \textit{\color{brown}Cross-Model experiments} the FR models used during the knowledge distillation step $M^t$ differs from the evaluation FR model, while in the \textit{\color{magenta}Same-Model experiments} the two are the same. When considering alignment, we separate the experiments into \textit{experiments with properly aligned images}, where the quality scores are predicted from optimally aligned faces samples for the targeted FR model, and \textit{experiments with misaligned images}, where the quality scores are extracted from face samples aligned with an arbitrary (non-optimal) landmark detector.

\vspace{1mm}\noindent\textbf{Experiments with Properly Aligned Images.} The results of the experiments with properly aligned images are shown in Table~\ref{tab:pauc_analysis_proper_align} for both the Cross- and Same-Model scenarios. Here, the average results across all datasets are marked as $\overline{pAUC}$. For each method, we show the baseline (B) results and the extended AI-KD approach (T), the better method of the two is marked with $*$ for individual datasets and with green for $\overline{pAUC}$. The best result of individual datasets is marked with \textbf{bold}. From the results, we observe that the proposed AI-KD  approach outperforms the baseline FIQA approaches, for all included FIQA techniques and on all tested FR models in terms of overall $\overline{pAUC}$ scores. Interestingly, the results on individual datasets are relatively close between the baseline and extended approaches, with the biggest improvements seen mostly on the hardest of the benchmarks XQLFW. 

\vspace{1mm}\noindent\textbf{Experiments with Misaligned Images.} From the results in  Table~\ref{tab:pauc_analysis_misalignment_same_model} for the Same-Model scenario and in Table~\ref{tab:pauc_analysis_misalignment_cross_model} for the Cross-Model scenario, we again see that for all combinations of benchmarks, FIQA techniques and FR models, the extended methods using AI-KD perform better than the baseline methods. One exception is when using CR-FIQA and DifFIQA(R) in combination with the SwinFace FR model in the Cross-Model scenario, where the baseline approach outperforms the extended approach by a slight margin. When looking at the results per individual benchmark the results appear to vary quite a bit between the different methods. The largest variation can be seen on the most difficult benchmark XQLFW, while for all others the differences between the extended and the baseline approaches appears to be significantly smaller. 

\subsection{Comparison with the State-of-the-Art}\label{sec:experiments_and_results:sota}

In this section, we compare the extended/distilled techniques to state-of-the-art FIQA methods, for both properly aligned and misaligned samples. The results using proper alignment are shown in Table~\ref{tab:pauc_sota_proper_alignment}, while the results using misaligned samples are reported in Table~\ref{tab:pauc_sota_proper_misalignment}. In both tables pAUC scores at a discard rate of $0.3$ and a FMR of $10^{-3}$ are shown. Average results across all datasets are marked as $\overline{pAUC}$, the best result on individual datasets are marked {bold}, the overall best result is marked {\color{green}green}, the second-best {\color{blue}blue} and the third-best {\color{red}red}.

\vspace{1mm}\noindent\textbf{Experiments with Aligned Images.} From the results in Table~\ref{tab:pauc_sota_proper_alignment}, we  observe that not only can the proposed knowledge distillation scheme improve the performance of existing FIQA techniques, it can easily achieve state-of-the-art results on all tested scenarios. The top performing method across the Cross- and Same-model scenarios is the extended DifFIQA(R) approach, achieving either the best or second-best result across all cases, closely followed by the extended CR-FIQA approach and FaceQAN. While the extended SER-FIQ outperforms the baseline, its results do not hold up against the top three contending methods. However, compared to all other tested methods, it still leads to competitive results.

\vspace{1mm}\noindent\textbf{Experiments with Misaligned Images.} For readability’s sake, the results in Table~\ref{tab:pauc_sota_proper_misalignment} with misaligned images are combined into a single score by averaging over all experiments. The results again tell a similar story, the best performing method is the extended DifFIQA(R) method, followed by the extended CR-FIQA and FaceQAN methods. However, here the divide appears to widen, as the extended methods achieve a marginally better result than for instance the third-best approach FaceQAN. With the extended SER-FIQ, we once again observe that it outperforms all remaining methods, except for the three front-runners. Overall, the results suggest that the alignment-invariant knowledge distillation not only improves performance when using misaligned samples, but is also beneficial for the performance of the FIQA techniques with properly aligned samples as well.

\begin{table}[!t!]
    \centering
    \caption{{Comparison with the state-of-the-art -- misaligned samples.} 
    }
    \renewcommand{\arraystretch}{1.05}
    \resizebox{\columnwidth}{!}{%
\begin{NiceTabularX}{\columnwidth}{l l | c c c c c c | r}

\toprule
 {\Block[fill=brown!10,l]{36-1}{\rotate \textbf{Cross-Model}}} & \multicolumn{8}{c}{\textbf{AdaFace}} \\
\cmidrule{2-9}
 & & \textbf{Adience} & \textbf{LFW} & \textbf{CPLFW} & \textbf{CFP-FP} & \textbf{CALFW} & \textbf{XQLFW} & $\overline{pAUC}$ \\ 
\cmidrule{2-9}
 & \textbf{FaceQAN}~\cite{faceqan} &  $0.910$ &  $0.760$ &  $0.767$ &  $\mathbf{0.622}$ &  $0.926$ &  $0.684$ & \cellcolor{magenta!10} $0.778$  \\
 & \textbf{SER-FIQ}~\cite{serfiq} &  $0.909$ &  $0.882$ &  $0.800$ &  $0.693$ &  $0.876$ &  $0.848$ &  $0.834$  \\
 & \textbf{FaceQnet}~\cite{faceqnet} &  $0.984$ &  $0.850$ &  $0.941$ &  $0.790$ &  $0.940$ &  $0.983$ &  $0.915$  \\
 & \textbf{SDD-FIQA}~\cite{sddfiqa} &  $0.909$ &  $0.882$ &  $0.800$ &  $0.693$ &  $0.876$ &  $0.848$ &  $0.834$  \\
 & \textbf{DifFIQA(R)}~\cite{diffiqar} &  $0.900$ &  $0.870$ &  $0.773$ &  $0.707$ &  $0.899$ &  $\mathbf{0.647}$ &  $0.799$  \\
 & \textbf{MagFace}~\cite{magface} &  $0.925$ &  $0.894$ &  $0.795$ &  $0.682$ &  $0.864$ &  $0.894$ &  $0.842$  \\
 & \textbf{CR-FIQA}~\cite{crfiqa} &  $0.912$ &  $0.832$ &  $0.742$ &  $0.676$ &  $\mathbf{0.858}$ &  $0.697$ &  $0.786$  \\
\cmidrule{2-9}
 & \textbf{AI-KD(SER-FIQ)} &  $0.884$ &  $0.894$ &  $0.800$ &  $0.702$ &  $0.900$ &  $0.719$ &  $0.816$  \\
 & \textbf{AI-KD(CR-FIQA)} &  $\mathbf{0.878}$ &  $0.792$ &  $\mathbf{0.734}$ &  $0.629$ &  $0.887$ &  $0.683$ & \cellcolor{cyan!10} $0.767$  \\
 & \textbf{AI-KD(DifFIQA(R))} &  $0.902$ &  $\mathbf{0.735}$ &  $0.736$ &  $0.645$ &  $0.885$ &  $0.671$ & \cellcolor{green!10} $0.762$  \\
\cmidrule{2-9}

 & \multicolumn{8}{c}{\textbf{SwinFace}} \\
\cmidrule{2-9}
 & & \textbf{Adience} & \textbf{LFW} & \textbf{CPLFW} & \textbf{CFP-FP} & \textbf{CALFW} & \textbf{XQLFW} & $\overline{pAUC}$ \\ 
\cmidrule{2-9}
 & \textbf{FaceQAN}~\cite{faceqan} &  $0.881$ &  $0.762$ &  $0.784$ &  $0.569$ &  $0.939$ &  $0.643$ &  $0.763$  \\
 & \textbf{SER-FIQ}~\cite{serfiq} &  $0.882$ &  $0.872$ &  $0.782$ &  $0.617$ &  $0.864$ &  $0.784$ &  $0.800$  \\
 & \textbf{FaceQnet}~\cite{faceqnet} &  $0.948$ &  $0.875$ &  $0.876$ &  $0.694$ &  $0.942$ &  $0.944$ &  $0.880$  \\
 & \textbf{SDD-FIQA}~\cite{sddfiqa} &  $0.882$ &  $0.872$ &  $0.782$ &  $0.617$ &  $0.864$ &  $0.784$ &  $0.800$  \\
 & \textbf{DifFIQA(R)}~\cite{diffiqar} &  $0.873$ &  $0.869$ &  $0.773$ &  $0.643$ &  $0.877$ &  $\mathbf{0.591}$ &  $0.771$  \\
 & \textbf{MagFace}~\cite{magface} &  $0.895$ &  $0.905$ &  $0.818$ &  $0.654$ &  $0.855$ &  $0.895$ &  $0.837$  \\
 & \textbf{CR-FIQA}~\cite{crfiqa} &  $0.875$ &  $0.836$ &  $\mathbf{0.757}$ &  $0.544$ &  $\mathbf{0.819}$ &  $0.661$ & \cellcolor{magenta!10} $0.749$  \\
\cmidrule{2-9}
 & \textbf{AI-KD(SER-FIQ)} &  $\mathbf{0.848}$ &  $0.895$ &  $0.808$ &  $0.623$ &  $0.835$ &  $0.658$ &  $0.778$  \\
 & \textbf{AI-KD(CR-FIQA)} &  $0.849$ &  $0.788$ &  $0.758$ &  $\mathbf{0.507}$ &  $0.887$ &  $0.640$ & \cellcolor{green!10} $0.738$  \\
 & \textbf{AI-KD(DifFIQA(R))} &  $0.871$ &  $\mathbf{0.761}$ &  $0.758$ &  $0.539$ &  $0.870$ &  $0.630$ & \cellcolor{cyan!10} $0.738$  \\
\cmidrule{2-9}

 & \multicolumn{8}{c}{\textbf{TransFace}} \\
\cmidrule{2-9}
 & & \textbf{Adience} & \textbf{LFW} & \textbf{CPLFW} & \textbf{CFP-FP} & \textbf{CALFW} & \textbf{XQLFW} & $\overline{pAUC}$ \\ 
\cmidrule{2-9}
 & \textbf{FaceQAN}~\cite{faceqan} &  $0.899$ &  $\mathbf{0.761}$ &  $0.740$ &  $0.622$ &  $0.960$ &  $0.636$ &  $0.769$  \\
 & \textbf{SER-FIQ}~\cite{serfiq} &  $0.904$ &  $0.882$ &  $0.792$ &  $0.718$ &  $0.906$ &  $0.800$ &  $0.834$  \\
 & \textbf{FaceQnet}~\cite{faceqnet} &  $0.960$ &  $0.847$ &  $0.937$ &  $0.823$ &  $0.967$ &  $0.998$ &  $0.922$  \\
 & \textbf{SDD-FIQA}~\cite{sddfiqa} &  $0.904$ &  $0.882$ &  $0.792$ &  $0.718$ &  $0.906$ &  $0.800$ &  $0.834$  \\
 & \textbf{DifFIQA(R)}~\cite{diffiqar} &  $0.889$ &  $0.879$ &  $0.741$ &  $0.726$ &  $0.907$ &  $\mathbf{0.565}$ &  $0.784$  \\
 & \textbf{MagFace}~\cite{magface} &  $0.927$ &  $0.893$ &  $0.780$ &  $0.690$ &  $0.897$ &  $0.918$ &  $0.851$  \\
 & \textbf{CR-FIQA}~\cite{crfiqa} &  $0.894$ &  $0.835$ &  $0.722$ &  $0.672$ &  $0.888$ &  $0.603$ & \cellcolor{magenta!10} $0.769$  \\
\cmidrule{2-9}
 & \textbf{AI-KD(SER-FIQ)} &  $\mathbf{0.854}$ &  $0.894$ &  $0.780$ &  $0.697$ &  $0.908$ &  $0.638$ &  $0.795$  \\
 & \textbf{AI-KD(CR-FIQA)} &  $0.867$ &  $0.792$ &  $0.703$ &  $\mathbf{0.601}$ &  $0.926$ &  $0.618$ & \cellcolor{cyan!10} $0.751$  \\
 & \textbf{AI-KD(DifFIQA(R))} &  $0.890$ &  $0.765$ &  $\mathbf{0.702}$ &  $0.627$ &  $\mathbf{0.886}$ &  $0.571$ & \cellcolor{green!10} $0.740$  \\

\bottomrule
\toprule

{\Block[fill=magenta!10,l]{12-1}{\rotate \textbf{Same-Model}}} & \multicolumn{8}{c}{\textbf{CosFace}} \\
\cmidrule{2-9}
 & & \textbf{Adience} & \textbf{LFW} & \textbf{CPLFW} & \textbf{CFP-FP} & \textbf{CALFW} & \textbf{XQLFW} & $\overline{pAUC}$ \\ 
\cmidrule{2-9}
 & \textbf{FaceQAN}~\cite{faceqan} &  $0.895$ &  $0.738$ &  $0.785$ &  $0.651$ &  $0.962$ &  $0.594$ & \cellcolor{magenta!10} $0.771$  \\
 & \textbf{SER-FIQ}~\cite{serfiq} &  $0.891$ &  $0.846$ &  $0.822$ &  $0.699$ &  $0.903$ &  $0.727$ &  $0.814$  \\
 & \textbf{FaceQnet}~\cite{faceqnet} &  $0.976$ &  $0.847$ &  $0.918$ &  $0.776$ &  $0.959$ &  $0.914$ &  $0.898$  \\
 & \textbf{SDD-FIQA}~\cite{sddfiqa} &  $0.891$ &  $0.846$ &  $0.822$ &  $0.699$ &  $0.903$ &  $0.727$ &  $0.814$  \\
 & \textbf{DifFIQA(R)}~\cite{diffiqar} &  $0.877$ &  $0.843$ &  $0.789$ &  $0.733$ &  $0.899$ &  $\mathbf{0.572}$ &  $0.785$  \\
 & \textbf{MagFace}~\cite{magface} &  $0.919$ &  $0.877$ &  $0.788$ &  $0.699$ &  $0.900$ &  $0.865$ &  $0.841$  \\
 & \textbf{CR-FIQA}~\cite{crfiqa} &  $0.899$ &  $0.816$ &  $0.770$ &  $0.658$ &  $\mathbf{0.889}$ &  $0.637$ &  $0.778$  \\
\cmidrule{2-9}
 & \textbf{AI-KD(SER-FIQ)} &  $0.872$ &  $0.891$ &  $0.822$ &  $0.716$ &  $0.909$ &  $0.604$ &  $0.802$  \\
 & \textbf{AI-KD(CR-FIQA)} &  $\mathbf{0.862}$ &  $0.783$ &  $0.757$ &  $\mathbf{0.616}$ &  $0.925$ &  $0.626$ & \cellcolor{cyan!10} $0.761$  \\
 & \textbf{AI-KD(DifFIQA(R))} &  $0.877$ &  $\mathbf{0.737}$ &  $\mathbf{0.756}$ &  $0.638$ &  $0.895$ &  $0.586$ & \cellcolor{green!10} $0.748$  \\

\bottomrule

    \end{NiceTabularX}
}
        \label{tab:pauc_sota_proper_misalignment}\vspace{-4mm}
\end{table}

\section{Conclusion}\label{sec:conclusion}

We presented a novel knowledge distillation technique, named AI-KD, which tries to improve the performance of existing FIQA methods on samples aligned with, from the viewpoint of the FIQA method, an unknown face landmark detector. Through extensive experiments, we showed that the proposed method is able to improve results not only on misaligned but also on properly aligned face images.

\scriptsize
\bibliographystyle{ieeetr}
\bibliography{bib}

\end{document}